\documentclass[10pt,twocolumn,letterpaper]{article}

\usepackage[pagenumbers]{cvpr} 

\usepackage{multirow}
\usepackage{graphicx}

\definecolor{cvprblue}{rgb}{0.21,0.49,0.74}
\usepackage[pagebackref,breaklinks,colorlinks,allcolors=cvprblue]{hyperref}
\usepackage{graphicx}
\usepackage{booktabs}
\usepackage{xcolor}
\usepackage{colortbl}

\definecolor{best}{HTML}{6cb1ff}   
\definecolor{second}{HTML}{c6e0fd}
\definecolor{third}{HTML}{ffffff}  

\graphicspath{{figures/}}
\def\confName{CVPR}
\def\confYear{2027}

\title{HazeFlux: Few-Step Generative Priors for Real-World Image Dehazing}

\author{
Yunong Liu$^{1,2}$\quad
Danna Xue$^{2}$\footnotemark[1]\quad 
Javier Vazquez-Corral$^{1,2}$\\
$^1$ Universitat Autònoma de Barcelona \quad $^2$ Computer Vision Center\\
{\tt\small \{yunong.liu, dxue, javier.vazquez\}@cvc.uab.cat}
}

\begin{document}
\twocolumn[{%
\renewcommand\twocolumn[1][]{#1}%
\maketitle

\begin{center}
    \centering
    \captionsetup{type=figure}
    \vspace{-2mm}
    \includegraphics[width=0.98\textwidth]{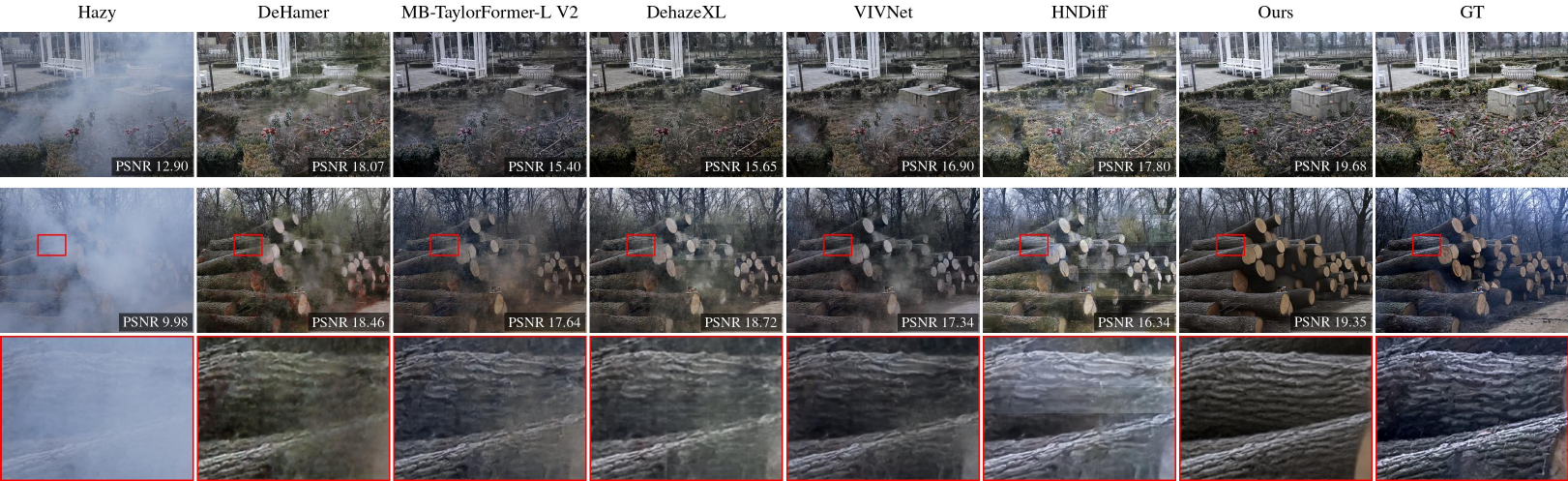}
\caption{Real image dehazing is a very challenging problem due to complex, non-homogeneous haze distributions and severe color degradation. Existing state-of-the-art methods, including DeHamer~\cite{guo2022image}, MB-TaylorFormer V2~\cite{jin2025mb}, DehazeXL~\cite{chen2025tokenize}, VIVNet~\cite{cui2026visual}, and HNDiff~\cite{he2026hndiff}, often fail to recover fine structural details or eliminate residual haze in heavy real-world scenarios. In contrast,
our method takes the advantages of the strong generative priors, so that it is able to restore intricate textures and faithful colors, producing results closest to the ground truth. }
\label{fig:teaser}
\end{center}%
}]

\renewcommand{\thefootnote}{\fnsymbol{footnote}}
\footnotetext[1]{Corresponding author.}

\begin{abstract}
Real-world single-image dehazing suffers from severe domain shifts, as paired real data is scarce and models trained on synthetic haze generalize poorly to real-world haze. While pre-trained generative editing models offer powerful image priors to bridge this generalization gap, applying them to image dehazing poses two major challenges: standard generative models prioritize distribution alignment over strict pixel-level fidelity, and operating in compressed latent spaces inherently discards fine-grained structural details. In this paper, we propose a novel two-stage framework that adapts the generative prior of a few-step flow-matching backbone to real-world dehazing. In the first stage, we fine-tune the latent backbone using Low-Rank Adaptation (LoRA), supervised not only on a flow-matching objective, but also on pixel-level supervision, thus enabling direct hazy-to-clear translation. In the second stage, we introduce structure-preserving skip connections directly to the VAE decoder, bypassing latent compression artifacts to recover precise spatial edges.  Experiments on four real-world benchmarks demonstrate that our approach significantly outperforms existing dehazing methods in real conditions. We commit to releasing the code upon acceptance.

\end{abstract}

\section{Introduction}
\label{sec:intro}

Haze degrades contrast, color fidelity, and overall visibility, severely impairing both human visual perception and downstream computer vision tasks in safety-critical applications, such as object detection in traffic-surveillance systems \cite{li2018benchmarking, huang2020dsnet}. Historically, single-image dehazing methods have relied on the Koschmieder atmospheric scattering model \cite{koschmieder1924theorie, narasimhan2002vision}, formulated as:
\begin{equation} \label{eq:haze}
I(x) = J(x)t(x) + A(1 - t(x)),
\end{equation}

\noindent where $I(x)$ is the resulting hazy image, $J(x)$ represents the clean, haze-free image, $A$ denotes the global atmospheric light, and $t(x) = \mathrm{e}^{-\beta d(x)}$ is the transmission map governed by scene depth $d(x)$ and atmospheric attenuation $\beta$. Under this formulation, recovering the clean image $J(x)$ fundamentally depends on accurately estimating $A$ and $t(x)$.

Traditional model-based approaches, such as Dark Channel Prior (DCP) \cite{he2009single} and Colorlines \cite{fattal2014dehaze}, rely on hand-crafted statistical priors to estimate the transmission map. However, these deterministic heuristics struggle to capture the complex, non-uniform characteristics of real-world atmospheric haze. Consequently, recent research has shifted toward learning direct distribution mappings between clear and hazy images, with diffusion-based generative models achieving promising results \cite{he2026hndiff, lan2025exploiting}. Nevertheless, because paired real-world training data is extremely difficult to acquire, most existing learning-based methods rely heavily on synthetic datasets \cite{li2018benchmarking}. As a result, models trained on synthetic distributions suffer from severe domain shift when deployed on real-world hazy scenes, resulting in a sharp performance drop.

 To overcome these limitations, recent insights demonstrate that pre-trained generative foundation models possess extensive prior knowledge of natural image distributions and understand haze as a high-level visual concept \cite{fei2023generative, yang2024unleashing}. Driven by this foundation, we propose a novel framework that leverages the generative power of current few-step editing backbones like FLUX.2 [klein]~\cite{flux2klein-2025} for image dehazing. However, adapting these models to dehazing is not straightforward and requires applying two crucial modifications. First, we implement distilled image-to-image translation with Low-Rank Adaptation (LoRA) fine-tuning \cite{hu2021lora}. Importantly, we do so by supervising the LoRA modules not only under a flow-matching objective, but also at the pixel level. Even though a critical bottleneck remains: operating within a compressed image latent space inherently causes the loss of fine-grained structural information required for precise pixel-level reconstruction, imposing a hard ceiling on fidelity regardless of how well the flow is learned. We therefore introduce gated image structure adapters to fuse high-resolution structural information from the hazy input with decoder features, compensating for latent compression while fully retaining the generative model’s real-world generalization capability.

Our main contributions are as follows:
\begin{enumerate}[leftmargin=*,noitemsep]
    \item We propose a simple and effective dehazing method based on a pre-trained latent flow matching model, which requires only a small amount of real haze training data and enables efficient haze removal.
    \item We introduce a gated image structure adapter to compensate for the loss of spatial details caused by VAE compression.
    \item We achieve robust dehazing performance with only four inference steps across various real-world haze datasets, covering both homogeneous and non-homogeneous haze, as well as indoor and outdoor scenes.
\end{enumerate}


\section{Related Work}

\subsection{Single Image Dehazing}

Traditional single-image dehazing methods rely on the atmospheric scattering model in \cref{eq:haze}, but recovering a haze-free image $J$ from a single hazy observation is fundamentally ill-posed. Early approaches used hand-crafted priors derived from natural image statistics, including the Dark Channel Prior~\cite{he2009single}, Color-Lines~\cite{fattal2014dehaze}, depth-saturation attenuation~\cite{zhu2015fast}, and non-local color clustering~\cite{berman2016nonlocal}. Other methods adopted variational and Retinex-duality formulations to recover image contrast through energy minimization~\cite{galdran2017fusion,galdran2018duality}. However, these approaches are sensitive to violations of their underlying assumptions, often resulting in color distortions and halo artifacts, \textit{e.g.}, in sky regions or under dense haze.

To address these limitations, research efforts shifted towards learning-based deep neural networks. Early methods, such as DehazeNet~\cite{cai2016dehazenet} and AOD-Net~\cite{li2017aod}, estimate the parameters of the atmospheric scattering model, while subsequent approaches directly learn hazy-to-clear mapping by feature fusion~\cite{qin2020ffa,cui2023focal} or transformers~\cite{guo2022image, qiu2023mbtaylorformer, jin2025mb}.
Recent approaches further improve real-world restoration through structural and color feature modeling~\cite{cui2024revitalizing,fang2025guided}, high-resolution patch processing~\cite{chen2025tokenize}, and unified restoration representations~\cite{cui2026visual}. To further improve generalization to real-world haze, PTTD~\cite{chen2024prompt} and BiLaLoRA~\cite{Zhang2026BiLaLoRA} explore prompt-guided adaptation and semantic-aware LoRA tuning, respectively, while IPC-Dehaze~\cite{fu2025iterative} exploits a pretrained VQGAN~\cite{esser2021taming} codebook prior for detail reconstruction. Despite these advances, deterministic regression-based methods often produce over-smooth results under severe degradation and struggle to recover realistic textures and fine details.

\begin{figure*}[t]
    \centering
    \includegraphics[width=\textwidth]{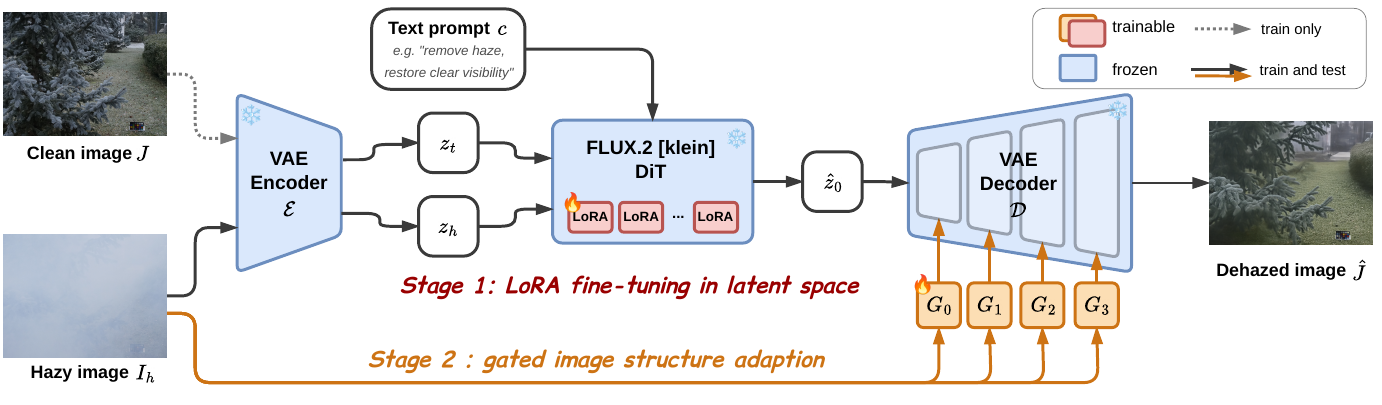}
    \caption{Overview of the proposed framework. In the first stage, we fine-tune the transformer in latent space, using the hazy latent $z_h$ as reference and a fixed prompt $c$, producing the clean latent estimate $\hat z_0$ in four steps with supervision from both latent space and pixel space. In the second stage, we compensate for the detail lost to latent compression: gated image structure adapters $G_0$–$G_3$ feed the hazy image into the four decoder levels under a learned per-pixel gate. The VAE and the FLUX.2 [klein]~\cite{flux2klein-2025} transformer stay frozen, and only the LoRA adapters and the gated image structure adapters are trained in the first and second stages, respectively. }
    \label{fig:overview}
    \vspace{-3mm}
\end{figure*}

\subsection{Diffusion Models in Image Dehazing}

Diffusion models~\cite{ho2020denoising,song2021scorebased} have emerged as powerful generative priors for image restoration. General frameworks such as Palette~\cite{saharia2022palette} and DDRM~\cite{kawar2022denoising} demonstrate the potential of diffusion priors for image-to-image translation and inverse problems, respectively. This paradigm has also been extended to adverse weather restoration~\cite{ozdenizci2023restoring} and unified multi-degradation restoration~\cite{zheng2024selective}.

Recent studies adapted diffusion models for real-world image dehazing. FCDM~\cite{wang2024frequency} combines frequency compensation with haze augmentation, while Diff-Dehazer~\cite{lan2025exploiting} exploits pre-trained diffusion priors alongside physical priors and image-text guidance. $\text{DiffLI}^2\text{D}$~\cite{yang2024unleashing} exploits semantic latent representations at different diffusion timesteps to disentangle image content from haze characteristics, thereby guiding the dehazing process.
DiffDehaze~\cite{hazegen2025} introduces a joint haze generation and removal framework, using HazeGen to synthesize training data. Both DehazeDDPM~\cite{yu2023high} and HNDiff~\cite{he2026hndiff} incorporates the atmospheric scattering model into the diffusion process, explicitly linking the generative process to the physical formation of haze. 

These methods demonstrate the potential of generative priors for recovering realistic image details in dehazing. However, conventional pixel-space diffusion models are computationally expensive, as their high-dimensional representations and iterative denoising processes typically require numerous model evaluations, hindering efficient high-resolution dehazing.

\subsection{Flow Matching and Latent Generative Models}

To reduce the computational burden of diffusion processes, Flow Matching (FM)~\cite{lipman2023flow} and Rectified Flow~\cite{liu2023flow} provide an alternative to conventional diffusion formulations by learning continuous vector fields that transport samples between probability distributions. In image dehazing, HazeFlow~\cite{shin2025hazeflow} reformulates the atmospheric scattering model as an ordinary differential equation (ODE) and learns a rectified-flow-inspired trajectory from hazy images to their clear counterparts, demonstrating promising reconstruction quality on real-world hazy images. Beyond conventional FM, Latent flow matching (LFM)~\cite{dao2023flow} further reduces the computational cost by performing generative modeling in the compact latent space of a pretrained autoencoder. This formulation benefits from the reduced dimensionality of latent representations and can leverage the generative priors of large-scale pretrained models. However, latent compression can discard spatial details, making it challenging to preserve fine structures during image restoration. This limitation is particularly relevant to dehazing, where high-frequency details may remain present in the hazy input even when they are difficult to perceive.

In this work, we adapt FLUX.2 [klein]~\cite{flux2klein-2025}, a state-of-the-art few-step image generation and editing model, for real-world image dehazing via a two-stage strategy. We first fine-tune the generative backbone using LoRA \cite{hu2021lora} for hazy-to-clear translation. We then introduce gated image structure adapters to convey high-resolution structural information from the hazy input to the VAE decoder, compensating for information loss caused by the latent bottleneck. This design combines the generative priors of the pretrained model with improved preservation of fine image structures.


\section{Methodology}

Given a hazy image $I_h\in\mathbb{R}^{3\times H\times W}$, our goal is to predict the corresponding clear image $\hat{J}\in\mathbb{R}^{3\times H\times W}$. We present HazeFlux, a latent flow-matching framework for single-image dehazing built on the pre-trained FLUX.2 [klein] model (see \cref{fig:overview}). We formulate dehazing as latent flow matching between the hazy and clear image distributions and adapt the pre-trained model in two stages. In Stage 1, we fine-tune the latent flow-matching Diffusion Transformer (DiT) using LoRA with supervision in both a flow-matching objective and at the pixel level to learn the mapping from hazy latents to clear latents. In Stage 2, we train a gated image structure adapter that leverages information from the hazy input $I_h$ to the VAE decoder in the pixel space to recover structural details lost during VAE compression.

\subsection{Preliminaries}
\subsubsection{Latent flow matching}

Let $\mathcal{E}$ and $\mathcal{D}$ denote the VAE encoder and decoder, respectively, and let $z_0=\mathcal{E}(J)$ denote the latent representation of the clear image. Flow matching defines a linear interpolation between Gaussian noise and data:
\begin{equation}
z_t=(1-t)\epsilon+tz_0,\quad \epsilon\sim\mathcal{N}(0,I),
\end{equation}
where $t\in[0,1]$ is the time variable. A velocity network $v_\theta$ is trained to predict the target velocity $z_0-\epsilon$ along this interpolation path by minimizing
\begin{equation}
\mathcal{L}_{\text{lat}}=
\mathbb{E}_{t,z_0,\epsilon}
\big\|v_\theta(z_t,t)-(z_0-\epsilon)\big\|_2^2,
\end{equation}
where $t$ is sampled uniformly from $[0,1]$. At inference, the learned velocity field defines an ODE, 
\begin{equation}
\frac{\mathrm{d}z_t}{\mathrm{d}t}=v_\theta(z_t,t),
\end{equation}
which transports a Gaussian noise sample at $t=0$ toward the data distribution at $t=1$. FLUX.2 [klein] is a step-distilled checkpoint based on this latent flow-matching formulation, enabling image generation with only a few function evaluations.

\subsubsection{Image-conditioned editing}

FLUX.2 [klein] performs image editing by conditioning generation on reference images, which are encoded into latent representations and packed into additional tokens concatenated with the noisy-latent tokens.

These tokens are jointly processed by the DiT, allowing the model to attend to both the reference image and the generated image. In image dehazing, the hazy input is considered as the reference. Let $z_h=\mathcal{E}(I_h)$ denote its latent representation, and let $c$ denote a fixed text prompt (\textit{e.g.}, ``remove haze, restore clear visibility''). The conditional velocity field is then given by $v_\theta(z_t,t,z_h,c)$. Note that the conditioning signal is itself an encoded latent, so the hazy observation reaches the DiT only after the same spatial compression applied to $z_0$.

\subsection{From Hazy Latent to Clean Latent}

In the first stage, the goal is to adapt the generative prior of FLUX.2 [klein], learned from large-scale natural images, to the hazy-to-clean mapping. We freeze the pre-trained transformer and the VAE, and insert rank-32 LoRA adapters into the image-stream attention projections of the double-stream blocks and the fused attention–MLP projections of the single-stream blocks. Given a paired sample $(I_h, J)$, we encode $z_0=\mathcal{E}(J)$ and $z_h=\mathcal{E}(I_h)$, draw $t$ and $\epsilon\sim\mathcal{N}(0,I)$, form $z_t=(1-t)\epsilon+t\,z_0$, and regress the velocity by:
\vspace{-2pt}
\begin{equation}
\mathcal{L}_{\text{lat}}=\mathbb{E}_{t,\epsilon}\big\|v_\theta(z_t,t,z_h,c)-(z_0-\epsilon)\big\|_2^2 .
\end{equation}
However, this latent objective places no constraint on the decoded output, and an error that is small in latent space can be amplified by the decoder to pixel space.
We therefore obtain an estimate of the clean latent by: $\hat z_0=z_t+(1-t)\,v_\theta(z_t,t,z_h,c)$, and decode it as $\hat J=\mathcal{D}(\hat z_0)$. We then apply pixel-level supervision:
\vspace{-2pt}
\begin{equation}
\mathcal{L}_{\text{pix}}=\lambda_1\|\hat J-J\|_1+\lambda_s\mathcal{L}_\mathrm{SSIM}(\hat J,J)+\lambda_p\,\mathcal{L}_\mathrm{LPIPS}(\hat J,J),
\end{equation}
\noindent where the $\mathcal{L}_1$ term constrains color and brightness, the $\mathcal{L}_\mathrm{SSIM}$ term (defined as 1 minus the SSIM index) constrains local structure, and the $\mathcal{L}_\mathrm{LPIPS}$ term~\cite{zhang2018unreasonable} constrains perceptual similarity.

$\mathcal{L}_1$ and $\mathcal{L}_\mathrm{SSIM}$ are applied only at timesteps sufficiently close to the target distribution ($t > \tau$), where $\hat z_0$ provides a meaningful estimate of the clean latent. $\mathcal{L}_\mathrm{LPIPS}$ uses a more restrictive threshold $\tau_p > \tau$, since $\hat J$ remains blurry at higher noise levels, and imposing perceptual similarity constraints on such outputs could encourage the model to hallucinate fine details. The overall objective of Stage 1 is:
\begin{equation}
\mathcal{L}_{\text{stage1}}=\mathcal{L}_{\text{lat}}+\mathcal{L}_{\text{pix}} .
\end{equation}
Since the VAE decoder remains frozen throughout this stage, pixel-level supervision can only improve the latent predictions produced by $v_\theta$, so the decoder's reconstruction limitations remain. This motivates Stage 2, which introduces high-resolution information from the hazy image to recover structural details that cannot be adequately reconstructed from the latent representation alone.

\begin{figure}[t]
    \centering
    \includegraphics[width=0.97\linewidth]{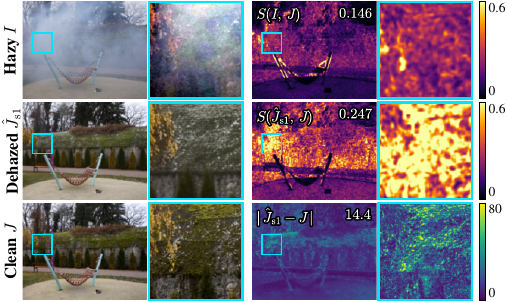}
    \caption{Pixel-wise $\ell_1$ differences and structural errors of the hazy input and Stage-1 output relative to the GT. \textbf{Row 1:} hazy input $I$ and its structural error map $S(I, J)$. \textbf{Row 2:} Stage-1 dehazed output $\hat{J}_\text{s1}$ and its structural error map $S(\hat{J}_\text{s1}, J)$. \textbf{Row 3:} ground truth $J$ and the pixel-wise $\ell_1$ difference map $|\hat{J}_\text{s1} - J|$. Here, $S(I_1,I_2)=1-s(I_1,I_2)$, where $s(I_1,I_2)$ denotes the structure term of SSIM~\cite{wang2004image} between images $I_1$ and $I_2$. Each error map is annotated with its spatial mean (top right). For visibility, only the zoomed-in hazy input is contrast-stretched, and this does not affect the computation of $S$.}
    \label{fig:skip_motivation}
    \vspace{-4mm}
\end{figure}

\subsection{Learning Structures from the Hazy Image}
\label{sec:stage2}

Although Stage~1 adapts the pre-trained model to haze removal, the reconstructed image $\mathcal{D}(\hat z_0)$ still lacks fine structural details due to information loss in the VAE latent representation. As shown in \cref{fig:skip_motivation}, the stage-1 output departs from the ground truth structurally more than the hazy input does, even though it is visually far cleaner. This indicates that fine details are lost in the latent space and not recovered faithfully by the VAE decoder. This motivates feeding pixel-level information from the hazy input into the decoder.

\begin{figure}[t]
    \centering
    \includegraphics[width=0.95\linewidth]{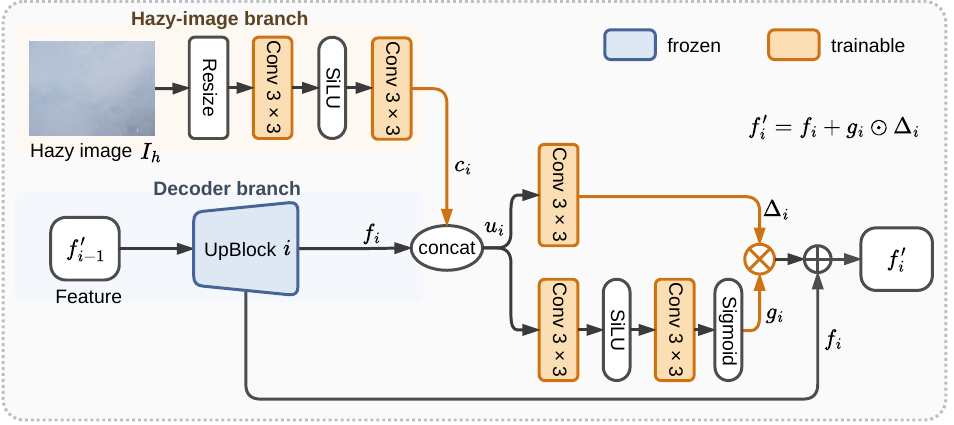}
    \caption{Architecture of the gated image structure adapter $G_i$ at decoder level $i$. Individual parameters are learned for each level.}
    \vspace{-3mm}
    \label{fig:skip}
\end{figure}

In the second stage, we introduce multi-level skip connections that feed pixel-level information from the hazy input into the VAE decoder (see \cref{fig:overview}). The decoder remains frozen, and the skip connections are trained to complement its latent reconstruction. We introduce a gated image structure adapter in each level of the VAE decoder, and its architecture is shown in \cref{fig:skip}. At each selected decoder level $i$, the hazy image $I_h$ is resized and passed through two $3 \times 3$ convolutions with an intermediate SiLU activation to obtain a conditioning feature $c_i$. This feature is concatenated with the corresponding decoder feature $f_i$ to form $u_i=[f_i;c_i]$. Two parallel heads then operate on $u_i$: a residual head predicts a correction $\Delta_i$ using a $3 \times 3$ convolution, while a gate head predicts a spatial gate $g_i$ that regulates the spatial contribution of the residual correction, using two $3 \times 3$ convolutions with an intermediate SiLU activation and a final sigmoid activation. The final fused feature is defined as:
\begin{equation}
f_i' = f_i + g_i \odot \Delta_i.
\end{equation}

Each decoder level has an independent adapter, allowing the network to adapt the injected information to different feature resolutions.

The hazy input provides structural information, such as edges and textures, that may be lost during latent encoding. However, it also contains haze-related degradations. The spatial gate therefore allows the network to selectively incorporate complementary structural information while suppressing potentially harmful corrections. This design preserves the latent path as the primary source of haze removal and uses the skip path to recover structural details that are difficult to reconstruct from the latent representation alone.

In the second stage, we keep the VAE and the DiT frozen, and train the gated image structure adapter using latents generated by the Stage~1 model to match the latent distribution encountered during inference. Given the generated latent $\hat z_0$ and the corresponding hazy image $I_h$, the detail-enhanced decoder produces $\hat J=\mathcal{D}_{\text{detail}}(\hat z_0;I_h)$. We supervise the dehazed image using pixel-level loss only, including the $\ell_1$ and the structural terms:
\begin{equation}
\mathcal{L}_\text{stage2}=\|\hat J-J\|_1+\lambda_s\mathcal{L}_\mathrm{SSIM}(\hat J,J),
\end{equation}
where $\lambda_s$ balances the fidelity and structural consistency.

\begin{table*}[t!]
\centering
\caption{Quantitative comparison with state-of-the-art methods on Dense-Haze, I-HAZE, O-HAZE, and NH-HAZE. \colorbox{best}{Best} and \colorbox{second}{second-best} models are marked.}
\label{tab:quantitative_comparison}
\resizebox{0.88\linewidth}{!}{
\setlength{\tabcolsep}{10pt}
\begin{tabular}{lcccccccc}
\toprule
\multirow{2}{*}{Method}
& \multicolumn{2}{c}{Dense-Haze}
& \multicolumn{2}{c}{I-HAZE}
& \multicolumn{2}{c}{O-HAZE}
& \multicolumn{2}{c}{NH-HAZE} \\
\cmidrule(lr){2-3}
\cmidrule(lr){4-5}
\cmidrule(lr){6-7}
\cmidrule(lr){8-9}
& PSNR & SSIM
& PSNR & SSIM
& PSNR & SSIM
& PSNR & SSIM \\
\midrule

DehazeDDPM [2023]~\cite{yu2023high}
& 15.45 & 0.456
& -- & --
& -- & --
& 19.44 & 0.628 \\

FocalNet [ICCV 2023]~\cite{cui2023focal}
& 16.95 & 0.597
& -- & --
& 25.46 & 0.791
& 20.36 & 0.696 \\

ConvIR [TPAMI 2024]~\cite{cui2024revitalizing}
& 16.86 & 0.600
& -- & --
& 25.25 & 0.784
& 20.65 & 0.692 \\

SGDN [AAAI 2025]~\cite{fang2025guided}
& 16.60 & 0.571
& -- & --
& 24.59 & 0.778
& 20.13 & 0.680 \\

DeHamer [CVPR 2022]~\cite{guo2022image}
& 16.62 & 0.560
& 19.44 & 0.790
& 22.42 & 0.693
& 20.66 & 0.684 \\

MB-TaylorFormer-B [ICCV 2023]~\cite{qiu2023mbtaylorformer}
& 16.66 & 0.560
& 22.10 & 0.781
& 25.05 & 0.788
& 20.43 & 0.688 \\

MB-TaylorFormer-L [ICCV 2023]~\cite{qiu2023mbtaylorformer}
& 16.44 & 0.566
& 21.09 & 0.762
& 25.31 & 0.782
& 20.49 & 0.692 \\

DehazeXL [CVPR 2025]~\cite{chen2025tokenize}
& 16.17 & 0.588
& 21.46 & \cellcolor{second}0.846
& 21.49 & 0.735
& 17.51 & 0.605 \\

HazeFlow [ICCV 2025]~\cite{shin2025hazeflow}
& 17.14 & 0.616
& -- & --
& -- & --
& 20.06 & 0.685 \\

MB-TaylorFormer-B V2 [TPAMI 2025]~\cite{jin2025mb}
& 16.95 & 0.621
& 22.29 & 0.777
& 25.29 & 0.790
& 20.73 & 0.703 \\

MB-TaylorFormer-L V2 [TPAMI 2025]~\cite{jin2025mb}
& 16.90 & 0.607
& \cellcolor{third}22.36 & 0.780
& \cellcolor{third}25.43 & 0.792
& 20.77 & \cellcolor{third}0.705 \\

VIVNet [TPAMI 2026]~\cite{cui2026visual}
& \cellcolor{second}17.26 & \cellcolor{best}0.632
& \cellcolor{second}23.00 & \cellcolor{third}0.811
& 24.74 & 0.780
& \cellcolor{third}21.01 & \cellcolor{second}0.706 \\

HNDiff [ECCV 2026]~\cite{he2026hndiff}
& \cellcolor{third}17.18 & \cellcolor{third}0.623
& -- & --
& \cellcolor{second}26.20 & \cellcolor{best}0.799
& \cellcolor{second}21.23 & 0.701 \\

\midrule
\textbf{HazeFlux (Ours)}
& \cellcolor{best}{18.01} & \cellcolor{second}{0.628}
& \cellcolor{best}{24.30} & \cellcolor{best}{0.854}
& \cellcolor{best}{26.46} & \cellcolor{second}{0.797}
& \cellcolor{best}{21.69} & \cellcolor{best}{0.714} \\

\bottomrule
\end{tabular}
}
\end{table*}

\begin{table*}[t]
\centering
\caption{Quantitative comparison with state-of-the-art methods on Merge-HAZE. \colorbox{best}{Best} and \colorbox{second}{second-best} models are marked. }
\label{tab:merge_haze_comparison}

\resizebox{0.88\linewidth}{!}{
\setlength{\tabcolsep}{6pt}
\begin{tabular}{lccccccccc}
\toprule
Method
& PSNR$\uparrow$
& SSIM$\uparrow$

& LPIPS$\downarrow$
& $\Delta E\downarrow$

& FID$\downarrow$
& Q-Align$\uparrow$
& RQI$\uparrow$ 
& CLIPIQA$\uparrow$
& TPIPS$\downarrow$\\
\midrule

DeHamer [CVPR 2022]~\cite{guo2022image}
& 18.72 & 0.676
& 0.524
& 11.91 & 
156.52
& 3.25 
& \cellcolor{second}0.464
& \cellcolor{best}0.46 
& 0.132\\

MB-TaylorFormer-B [ICCV 2023]~\cite{qiu2023mbtaylorformer}
& 19.52 & 0.631

& 0.498
& 10.65 

& 154.22
& 3.23 
& 0.443 
& 0.37
& 0.128\\

MB-TaylorFormer-L [ICCV 2023]~\cite{qiu2023mbtaylorformer}
& 18.66 & 0.628
& 0.507
& 12.11 
& 191.91
& 3.15 
& 0.440
& \cellcolor{third}0.41
& 0.125\\

MB-TaylorFormer-B V2 [TPAMI 2025]~\cite{jin2025mb}
& \cellcolor{second}20.37 & 0.679
& \cellcolor{third}0.480
& 9.98 
& \cellcolor{third}133.14
& \cellcolor{third}3.47 
& \cellcolor{third}0.458
& 0.38
& 0.106\\

MB-TaylorFormer-L V2 [TPAMI 2025]~\cite{jin2025mb}
& \cellcolor{third}20.15 & 0.658
& 0.488
& 10.29 
& 139.18
& 3.41 
& 0.457
& 0.37 
& 0.110\\

DehazeXL [CVPR 2025]~\cite{chen2025tokenize}
& 19.19 & \cellcolor{second}0.712
& 0.488
& \cellcolor{third}9.46 
& \cellcolor{second}129.66
& 3.44 
& 0.449
& 0.30
& \cellcolor{third}0.104\\

VIVNet [TPAMI 2026]~\cite{cui2026visual}
& 20.12 & 0.686
& \cellcolor{third}0.475
& \cellcolor{second}9.43 
& 133.88
& \cellcolor{second}3.60 
& 0.455
& 0.37
& \cellcolor{second}0.098\\

HNDiff [ECCV 2026] ~\cite{he2026hndiff}
& 18.75 & \cellcolor{third}0.707
& 0.491
& 10.72
& 145.02
& 3.17
& 0.451
& 0.30
& 0.114\\

\midrule
\textbf{HazeFlux (Ours)}
& \cellcolor{best}20.98 & \cellcolor{best}0.722
& \cellcolor{best}0.469
& \cellcolor{best}8.49
& \cellcolor{best}77.90
& \cellcolor{best}4.02 
& \cellcolor{best}0.489
& \cellcolor{second}0.44
& \cellcolor{best}0.056\\ 

\bottomrule
\end{tabular}
}
\vspace{-3mm}
\end{table*}

\subsection{Inference}
\label{sec:infer}
At inference, the hazy image is encoded into $z_h$. Starting from Gaussian noise, we generate the clear latent $\hat z_0$ through four flow-matching sampling steps, conditioned on $z_h$ and the prompt $c$, using a guidance scale of 1. The final dehazed image is then reconstructed by the skip-enhanced decoder: $\hat J=\mathcal{D}_{\text{detail}}(\hat z_0;I_h)$ (Fig.~\ref{fig:skip}). The pipeline thus consists of four sampling steps and a single decoding pass. We use the same sampling configuration to generate training latents for Stage~2 training.


\section{Results}

\subsection{Experimental Settings}

\noindent \textbf{Implementation details.} For stage 1, we fine-tune FLUX.2 [klein] 4B~\cite{flux2klein-2025} with LoRA of rank 32 on $512 \times 512$ patches, augmented with random horizontal flips. We train it for 3,000 steps with a batch size of 8, using 8-bit AdamW, a learning rate of $10^{-4}$ that is constant after 50 warm-up steps. For the pixel-level loss, we set $\lambda_1=10$, $\lambda_s=1$, and $\lambda_p=1$, with thresholds $\tau=0.3$ and $\tau_p=0.5$. The LPIPS loss is computed on the features from a VGG backbone, and the $\ell_1$ term is computed on images in $[-1,1]$. For stage 2 training, we use AdamW with a learning rate of $5\times10^{-5}$ without weight decay, 50 warm-up steps followed by cosine decay to 5\% of the initial learning rate, and 1,500 optimizer steps. We set $\lambda_s=1$, and obtain the training latents by running the stage-1 model with six different random seeds for each training image. The VAE and the skip-connection modules are run in FP32. The training is conducted on a single NVIDIA RTX PRO 6000 Blackwell GPU.

\noindent \textbf{Datasets.} We evaluate our method on four real-world dehazing benchmarks: Dense-Haze~\cite{ancuti2019dense}, NH-HAZE~\cite{ancuti2021ntire}, O-HAZE~\cite{ancuti2018haze}, and I-HAZE~\cite{ancuti2018haze}. These datasets cover different haze conditions and capture scenarios, including dense homogeneous haze, non-homogeneous haze, and both indoor and outdoor scenes. We first conduct standard evaluations on each dataset using its official training and test splits. To further assess model performance across diverse real-world conditions, we combine the four datasets into a unified dataset, which we refer to as Merge-HAZE. For this setting, we separately aggregate the training and test splits of all four datasets.

\noindent \textbf{Metrics.} For the standard per-dataset evaluations, we follow the protocols and measure performance using Peak Signal-to-Noise Ratio (PSNR) and Structural Similarity Index (SSIM)~\cite{wang2004image}. For the combined dataset setting, as all models are retrained, we extend our evaluation to incorporate color fidelity ($\Delta E$), perceptual similarity (LPIPS~\cite{zhang2018unreasonable} and TPIPS~\cite{TPIPS}), distribution realism (FID~\cite{fid}), perceptual visual quality via Q-Align~\cite{qalign} and RQI~\cite{su2026bridging}, and the no-reference metric of CLIP-IQA~\cite{wang2023clipiqa}.

\noindent \textbf{Competing methods.}
We compare our approach with recent state-of-the-art image dehazing methods, categorized into deterministic and generative approaches. Deterministic methods directly predict a haze-free image through feed-forward networks. This category includes DeHamer~\cite{guo2022image}, FocalNet~\cite{cui2023focal}, MB-TaylorFormer-B and MB-TaylorFormer-L~\cite{qiu2023mbtaylorformer}, ConvIR~\cite{cui2024revitalizing}, SGDN~\cite{fang2025guided}, DehazeXL~\cite{chen2025tokenize}, MB-TaylorFormer-B V2 and MB-TaylorFormer-L V2~\cite{jin2025mb}, and VIVNet~\cite{cui2026visual}. For generative methods, we include DehazeDDPM~\cite{yu2023high}, HNDiff~\cite{he2026hndiff}, and HazeFlow~\cite{shin2025hazeflow}.

\subsection{Quantitative Results} 

\noindent \textbf{Individual datasets.} \Cref{tab:quantitative_comparison} presents the quantitative results for the standard per-dataset training setting across both PSNR and SSIM. Our method ranks first in 6 out of 8 metrics, yielding notable gains in PSNR such as +1.30~dB on I-HAZE, +0.75~dB on Dense-Haze, and +0.46~dB on NH-HAZE. While VIVNet~\cite{cui2026visual} and HNDiff~\cite{he2026hndiff} are the closest competing baselines, our model retains a competitive second-place ranking in the remaining two cases (Dense-Haze SSIM, O-HAZE SSIM), trailing the best-performing model by less than 0.004 in SSIM.

\noindent \textbf{Merged dataset.} \Cref{tab:merge_haze_comparison} shows the quantitative results on the Merge-HAZE dataset. For this experiment, we compare our approach against those methods with better behaviour in the individual datasets, namely DeHamer~\cite{guo2022image}, the MB-TaylorFormer series ~\cite{qiu2023mbtaylorformer, jin2025mb}, DehazeXL~\cite{chen2025tokenize}, VIVNet~\cite{cui2026visual}, and HNDiff \cite{he2026hndiff}. We retrain all methods from scratch following the training protocols described in their original papers. In addition to PSNR and SSIM, we extend the evaluation to include LPIPS, TPIPS (with the prompt ``Dehazed image''), $\Delta E$, FID, Q-Align, RQI, and CLIP-IQA, providing a more comprehensive assessment of perceptual similarity, color fidelity, and distributional quality. Our method outperforms all competing methods across all evaluated metrics (except CLIP-IQA, where it ranks second), surpassing the closest competing model by +0.61~dB in PSNR and reducing color distortion by 0.94 $\Delta E$ units.

\noindent \textbf{Cross-dataset generalization.} 
To evaluate the generalization ability of models trained on limited real-world haze data, we test each model on the three datasets excluded from its training set. \Cref{tab:cross_dataset} reports the results of the three best-performing methods in \Cref{tab:quantitative_comparison}, namely, our method, HNDiff, and VIVNet, under this cross-dataset setting. Our method outperforms the other two methods in 20 out of 24 cases, suggesting that adapting a pre-trained image-editing backbone can improve generalization to unseen haze conditions. We also observe that models trained on Dense-Haze and O-HAZE generally achieve better cross-dataset performance than those trained on NH-HAZE and I-HAZE. This may be attributed to
differences in scene domains and haze distributions across datasets. I-HAZE is restricted to indoor environments, limiting its transferability to outdoor scenes, whereas NH-HAZE features spatially non-homogeneous haze that can cause models to under-correct globally uniform haze. In contrast, both Dense-Haze and O-HAZE combine outdoor scenes with homogeneous haze, and Dense-Haze further introduces substantially heavier haze density.

\begin{table}[t]
\caption{Cross-dataset generalization. Each block trains on one dataset
and tests on the remaining three. Best result per column within a block
in \textbf{bold}.}
\vspace{-2mm}
\label{tab:cross_dataset}

\centering
\scriptsize
\setlength{\tabcolsep}{2pt}
\renewcommand{\arraystretch}{1.1}
\begin{tabular}{llcccccccc}
\toprule
\multirow{2}{*}{\shortstack[c]{Train\\Data}}
& \multirow{2}{*}{Method}
& \multicolumn{2}{c}{\shortstack{Dense-Haze}}
& \multicolumn{2}{c}{I-HAZE}
& \multicolumn{2}{c}{O-HAZE}
& \multicolumn{2}{c}{NH-HAZE} \\
\cmidrule(lr){3-4}
\cmidrule(lr){5-6}
\cmidrule(lr){7-8}
\cmidrule(lr){9-10}
& & PSNR & SSIM
& PSNR & SSIM
& PSNR & SSIM
& PSNR & SSIM \\
\midrule

\multirow{3}{*}{\shortstack[l]{Dense-\\Haze}}
& VIVNet
& -- & --
& 19.68 & 0.750
& 18.32 & 0.599
& 15.20 & 0.535 \\

& HNDiff
& -- & --
& 16.02 & 0.722
& 18.12 & 0.593
& 16.44 & 0.624 \\

& Ours
& -- & --
& \textbf{20.23} & \textbf{0.823}
& \textbf{21.98} & \textbf{0.709}
& \textbf{19.23} & \textbf{0.649} \\

\midrule

\multirow{2}{*}{I-HAZE}
& VIVNet
& 13.85 & 0.514
& -- & --
& \textbf{21.68} & \textbf{0.695}
& 15.42 & \textbf{0.586} \\

& Ours
& \textbf{15.56} & \textbf{0.591}
& -- & --
& 20.00 & 0.684
& \textbf{15.80} & 0.574 \\

\midrule

\multirow{3}{*}{O-HAZE}
& VIVNet
& 14.41 & 0.549
& 18.52 & 0.795
& -- & --
& 15.98 & 0.547 \\

& HNDiff
& 13.15 & 0.559
& 17.81 & 0.804
& -- & --
& 16.47 & 0.558 \\

& Ours
& \textbf{17.30} & \textbf{0.623}
& \textbf{22.12} & \textbf{0.849}
& -- & --
& \textbf{18.49} & \textbf{0.660} \\

\midrule

\multirow{3}{*}{NH-HAZE}
& VIVNet
& 13.85 & 0.517
& 15.56 & 0.764
& 19.29 & 0.708
& -- & -- \\

& HNDiff
& \textbf{15.48} & 0.520
& 15.28 & 0.623
& 19.20 & 0.646
& -- & -- \\

& Ours
& 15.11 & \textbf{0.601}
& \textbf{18.79} & \textbf{0.818}
& \textbf{19.90} & \textbf{0.739}
& -- & -- \\

\bottomrule
\end{tabular}
\vspace{-2mm}
\end{table}

\begin{table}[t!]
\centering
\caption{Ablation of two-stage training on five haze datasets.}
\label{tab:skip_comparison}

\renewcommand{\arraystretch}{1}
\setlength{\tabcolsep}{3pt}

\resizebox{\linewidth}{!}{
\begin{tabular}{lccccccccccc}
\toprule

\multirow{2}{*}{Method}
& \multicolumn{2}{c}{Dense-Haze}
& \multicolumn{2}{c}{I-HAZE}
& \multicolumn{2}{c}{O-HAZE}
& \multicolumn{2}{c}{NH-HAZE}
& \multicolumn{3}{c}{Merge-HAZE} \\

\cmidrule(lr){2-3}
\cmidrule(lr){4-5}
\cmidrule(lr){6-7}
\cmidrule(lr){8-9}
\cmidrule(lr){10-12}

& PSNR & SSIM
& PSNR & SSIM
& PSNR & SSIM
& PSNR & SSIM 
& PSNR & SSIM & $\Delta E$ \\

\midrule
FLUX.2 [klein] 
& 10.53 & 0.506
& 10.20 & 0.599
& 11.88 & 0.400
& 14.96 & 0.519
& 11.94 & 0.506 & 22.317 \\

Ours (Stage 1) 
& 17.66 & 0.611
& \textbf{24.55} & 0.852
& 25.31& 0.730
& 21.15 & 0.651 
& 20.61 & 0.699 & 8.734 \\

Ours (Stage 2) 
& \textbf{18.01} & \textbf{0.628}
& 24.30 & \textbf{0.854}
&  \textbf{26.46} & \textbf{0.797}
& \textbf{21.69} & \textbf{0.714} 
& \textbf{20.98} & \textbf{0.722} & \textbf{8.488} \\

\bottomrule
\end{tabular}
}
\vspace{-1mm}
\end{table}

\begin{table}[t!]
\centering
\caption{Ablation on pixel losses (Stage 1) on I-HAZE.}
\label{tab:ablation}

\small
\setlength{\tabcolsep}{2.5pt}
\renewcommand{\arraystretch}{1}

\resizebox{\linewidth}{!}{
\begin{tabular}{lccccccccc}
\toprule
Loss
& PSNR$\uparrow$
& SSIM$\uparrow$
& LPIPS$\downarrow$
& $\Delta E\downarrow$
& FID$\downarrow$
& Q-Align$\uparrow$ \\
\midrule

FLUX.2 [klein] w/o finetune
& 10.20 
& 0.599
& 0.552
& 24.68 
& 433.84
& 1.98 \\ \midrule

$\mathcal{L}_{\text{lat}}$
& 23.62
& 0.839 
& 0.438
& 5.96
& 65.02
& 4.42 \\

$\mathcal{L}_{\text{lat}}+\mathcal{L}_1$
& 23.63
& 0.842
& 0.440
& 5.60
& 54.42
& \textbf{4.47}\\

$\mathcal{L}_{\text{lat}}+\mathcal{L}_1+\mathcal{L}_{\text{LPIPS}}$
& 24.08
& 0.851
& \textbf{0.436}
& \textbf{5.20}
& 59.24
& 4.43\\

$\mathcal{L}_{\text{lat}}+\mathcal{L}_1+\mathcal{L}_{\text{LPIPS}} +  \mathcal{L}_{\text{SSIM}}$
& \textbf{24.55}
& \textbf{0.852}
& 0.463
& 5.28
& \textbf{51.61}
& 4.46\\

\bottomrule
\end{tabular}
}
\vspace{-2mm}
\end{table}

\begin{figure}[t]
  \centering
  \includegraphics[width=0.95\columnwidth]{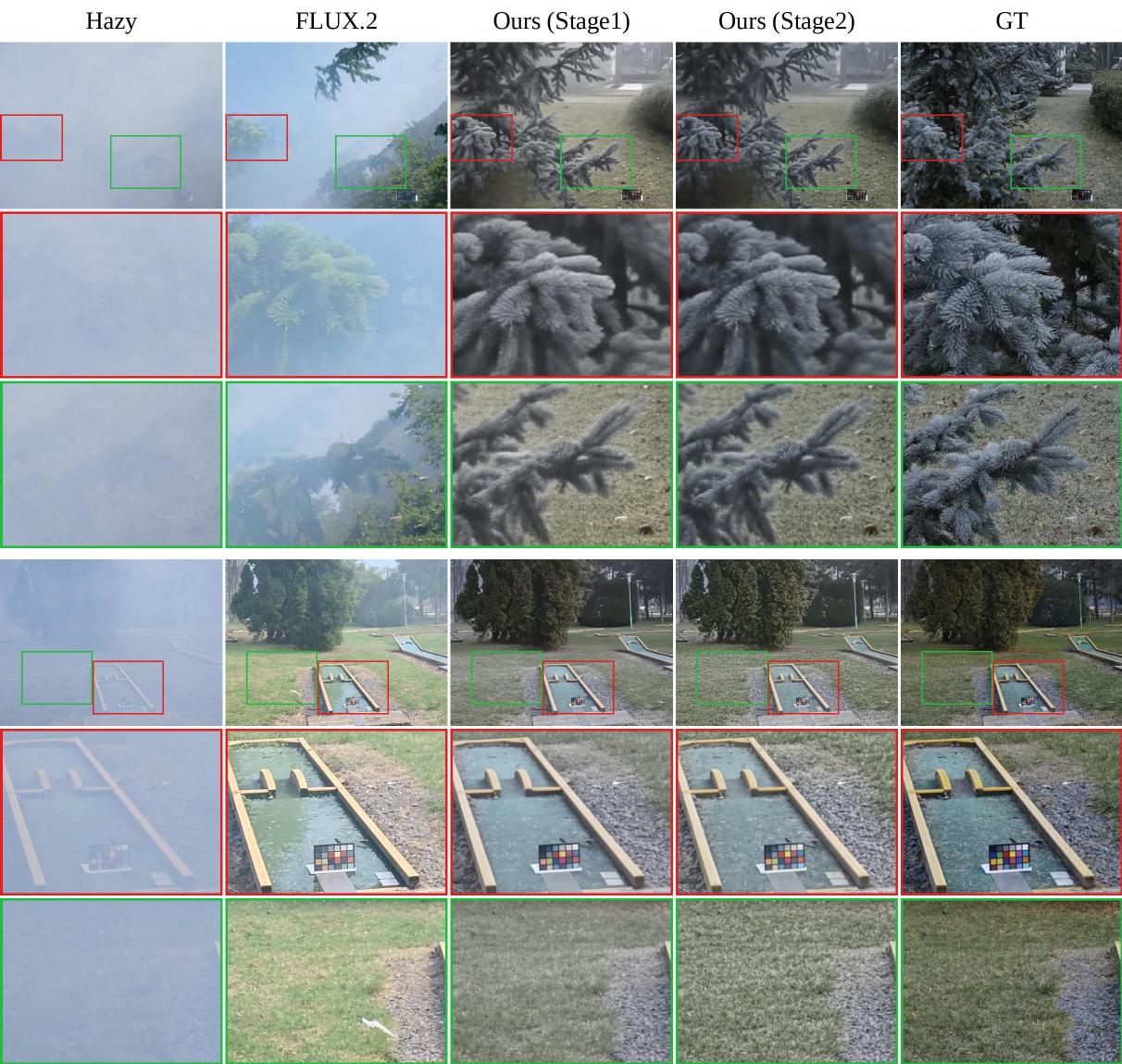}
  \caption{Comparison of the reconstruction details between the baseline FLUX.2 [klein], ours (stage 1), and ours (stage 2).}
  \label{fig:details}
  \vspace{-1mm}
\end{figure}

 \begin{figure}[t]
     \centering
     \includegraphics[width=0.95\linewidth]{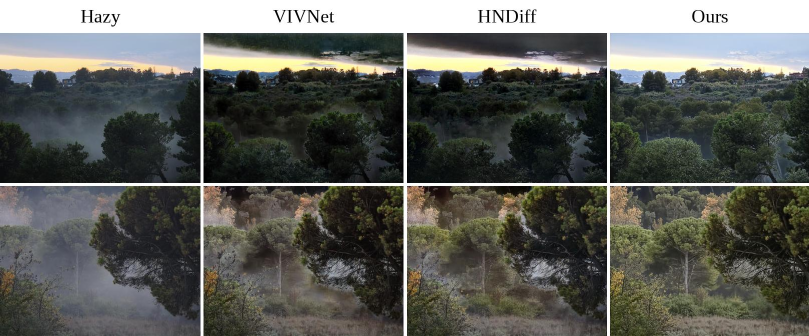}
     \caption{Qualitative comparison on real-world hazy images.}
     \label{fig:real}
     \vspace{-2mm}
 \end{figure}

\subsection{Ablation Study}

\noindent \textbf{Two-stage training paradigm.} Table~\ref{tab:skip_comparison} compares the original FLUX.2 [klein] model, our model after Stage 1, and the full model after both training stages. Stage 1 already brings substantial improvements over the original model, while Stage 2 provides further refinements. Notably, Stage 2 improves not only SSIM but also PSNR on all datasets except I-HAZE. On Merge-HAZE, the $\Delta E$ results also show progressive improvements from Stage 1 to Stage 2, indicating increasingly accurate color restoration. \Cref{fig:details} provides a qualitative comparison of the three configurations, with the original FLUX.2 [klein] model, Stage 1, and the full model shown from left to right. We highlight the leaves of the tree in the top row and the grass in the bottom row. The original model partially removes haze, while Stage 1 produces visibly cleaner results. Stage 2 further recovers fine details in both regions, demonstrating the complementary contributions of the two training stages.

\begin{figure*}[t]
    \centering
    \includegraphics[width=\textwidth]{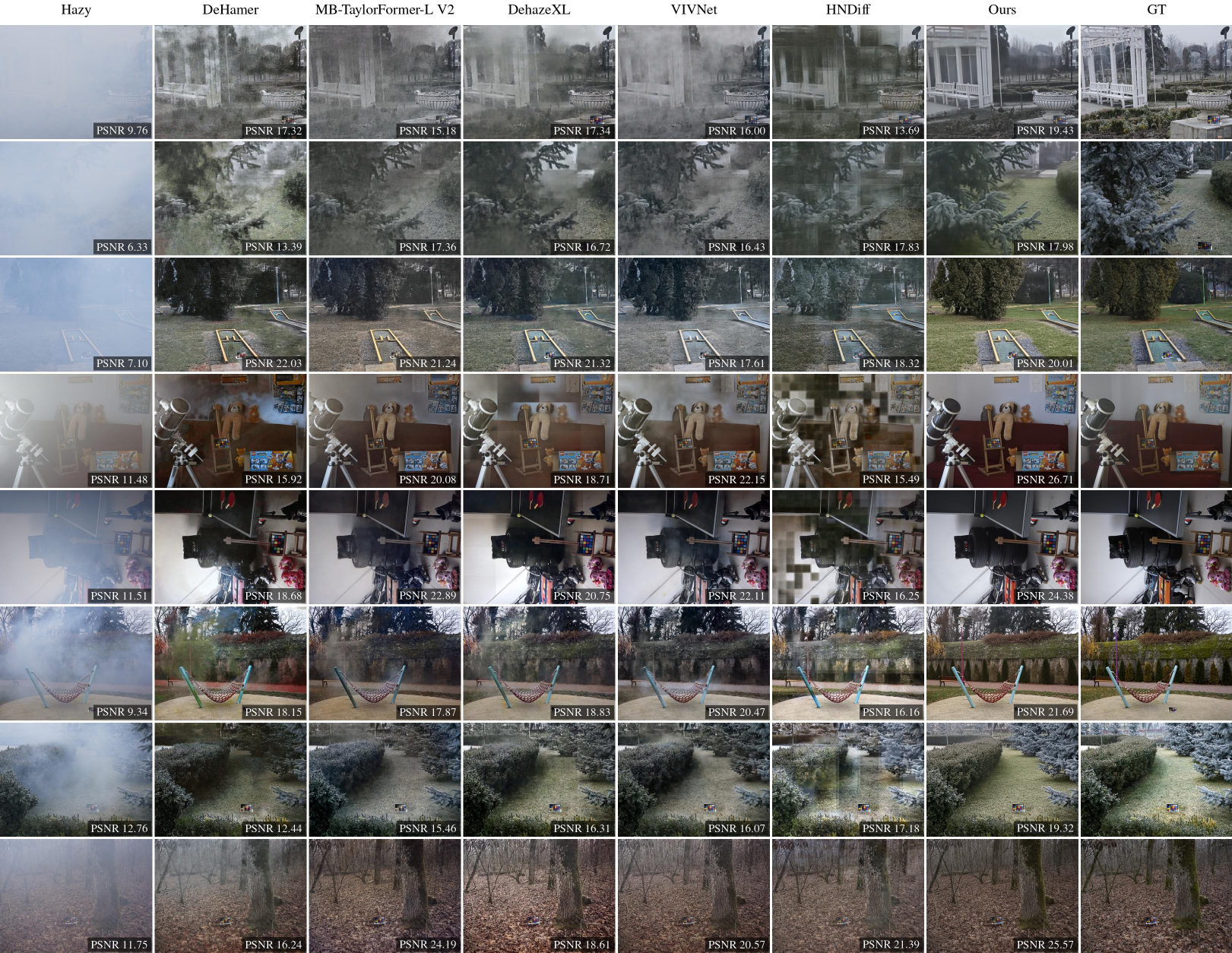}
    \vspace{-5mm}
    \caption{Qualitative comparison on hazy images from the I-HAZE, O-HAZE, Dense-Haze, and NH-HAZE datasets, including challenging cases such as severe haze (Rows 1 to 3) and non-homogeneous haze (Rows 6 and 7). From left to right, we show the original images, the results of DeHamer~\cite{guo2022image}, the MB-TaylorFormer V2~\cite{jin2025mb}, DehazeXL~\cite{chen2025tokenize}, VIVNet~\cite{cui2026visual}, and HNDiff \cite{he2026hndiff}, our results, and the ground truth.}
    \label{fig:comparison}
    \vspace{-2mm}
\end{figure*}

\noindent \textbf{Pixel-level supervision.} 
We conduct an ablation study on the pixel-level losses introduced during Stage 1. As shown in Table~\ref{tab:ablation}, compared with the model before fine-tuning, the latent loss brings the most substantial improvements, providing a strong baseline for subsequent pixel-level supervision. Adding the $\ell_1$ loss to the original latent loss leads to modest overall improvements, with the most notable gain in FID. Incorporating the perceptual LPIPS loss improves most evaluation metrics, except FID and Q-Align. Adding SSIM supervision obtains the best results for PSNR, SSIM, and FID, and improves Q-Align over the previous step, only minimally trailing on $\Delta E$ and the expected decay in LPIPS. 

\subsection{Qualitative Results} 
\cref{fig:real} and \cref{fig:comparison} present qualitative comparisons between our method trained on Merge-HAZE and state-of-the-art dehazing methods. Most competing methods leave noticeable residual haze in challenging conditions in \cref{fig:comparison}, including severe haze (Rows 1--3) and non-homogeneous haze (Rows 6--7). 
In contrast, our method produces visually cleaner, clearer results across indoor and outdoor scenes while better preserving fine details and avoiding the tile-boundary artifacts seen in DehazeXL and HNDiff.
\cref{fig:real} further demonstrates the generalization ability of our method on two real-world hazy images captured with our smartphone. Our method removes haze more effectively while avoiding artifacts in the highlighted regions of the sky.

\section{Conclusion}
In this paper, we presented a two-stage framework to adapt pre-trained generative editing models for real-world image dehazing. Our first stage fine-tunes the latent backbone using LoRA under a joint objective combining flow matching with pixel-level supervision. To overcome spatial detail loss caused by latent compression, our second stage introduces gated image structure adapters to the VAE decoder, further recovering sharp edges and intricate textures. Experiments across four real-world datasets (Dense-Haze, NH-HAZE, O-HAZE, and I-HAZE) and the combined Merge-HAZE dataset validate the effectiveness of our approach, outperforming existing state-of-the-art methods in both quantitative metrics and visual quality.

\noindent \textbf{Limitations and future work.} Despite its strong performance, our approach inherits certain characteristics from the underlying FLUX.2 [klein] editing backbone. Specifically, because the backbone was originally trained for creative image editing, the generative prior can occasionally produce slight color over-saturation in complex, heavily hazed regions. In future work, we plan to explore adaptive color-constraining objectives to regulate saturation while preserving the model's high-fidelity structural restoration.

\clearpage
\section*{LLM Usage Statement}

We used ChatGPT and Claude in preparing this work.

\noindent \textbf{Writing:} The tools were used for grammar correction, rewriting and polishing the phrasing of the manuscript. All technical claims, experimental descriptions and conclusions were written and verified by the authors.

\noindent \textbf{Code:} The tools assisted with auxiliary scripts, specifically
figure plotting and result tabulation. The method implementation, training and evaluation protocol were written by the authors.

\noindent \textbf{Literature:} The tools were used to search for related literature. Every cited paper was read and verified by the authors.

\noindent No part of the proposed method, experimental design or analysis was generated by an LLM. 

\noindent The authors take full responsibility for the
content of this paper.

{
    \small
    \bibliographystyle{ieeenat_fullname}
    \bibliography{main}
}
\clearpage
\setcounter{page}{1}
\setcounter{section}{0}
\renewcommand{\thesection}{\Alph{section}}
\maketitlesupplementary
In this supplementary material, we provide an additional ablation study on the LoRA rank used in Stage 1 in \cref{sup_sec:ablation}, a quantitative comparison with BiLaLoRA \cite{Zhang2026BiLaLoRA} in \cref{sup_sec:bilalora},
additional qualitative comparisons on real hazy images in \cref{sup_sec:quality}, and details about our computational cost in \cref{sup_sec:efficiency}.

\section{Additional Ablation on LoRA Rank}
\label{sup_sec:ablation}
We study the effect of the LoRA rank in Stage 1 on I-HAZE in Table~\ref{tab:ablation of rank}. 

Increasing the rank from 8 to 32 improves most metrics, with the exception of LPIPS. Rank 16 performs similarly to rank 32 on $\Delta E$ and the Q-Align score, but we kept rank 32 because of the better PSNR and its modest cost (12.55M trainable parameters; Table~\ref{tab:params-speed}). Going further up to rank 64 would make the model excessively heavy, so we did not experiment with it.

\section{Comparison with BiLaLoRA}
\label{sup_sec:bilalora}
BiLaLoRA \cite{Zhang2026BiLaLoRA} adapts a dehazing network pre-trained on synthetic haze to real haze. We compare with BiLaLoRA under two settings in Table~\ref{tab:bilalora comparison}. 

In the first setting, BiLaLoRA (fine-tuned), we start from its first-stage weights and fine-tune them on the training split of each of the four datasets (Dense-Haze, I-HAZE, O-HAZE, and NH-HAZE) using its own unsupervised CLIP loss, then evaluate on the corresponding test split. This gives BiLaLoRA the same in-domain training data that our method uses.

In the second setting, BiLaLoRA (released), we directly evaluate the weights obtained after its own fine-tuning, without any further training on the benchmark datasets.

Our method clearly improves BiLaLoRA on all the datasets.

\section{Additional Qualitative Results}
\label{sup_sec:quality}

\begin{figure*}[t!]
  \centering
  \includegraphics[width=\textwidth]{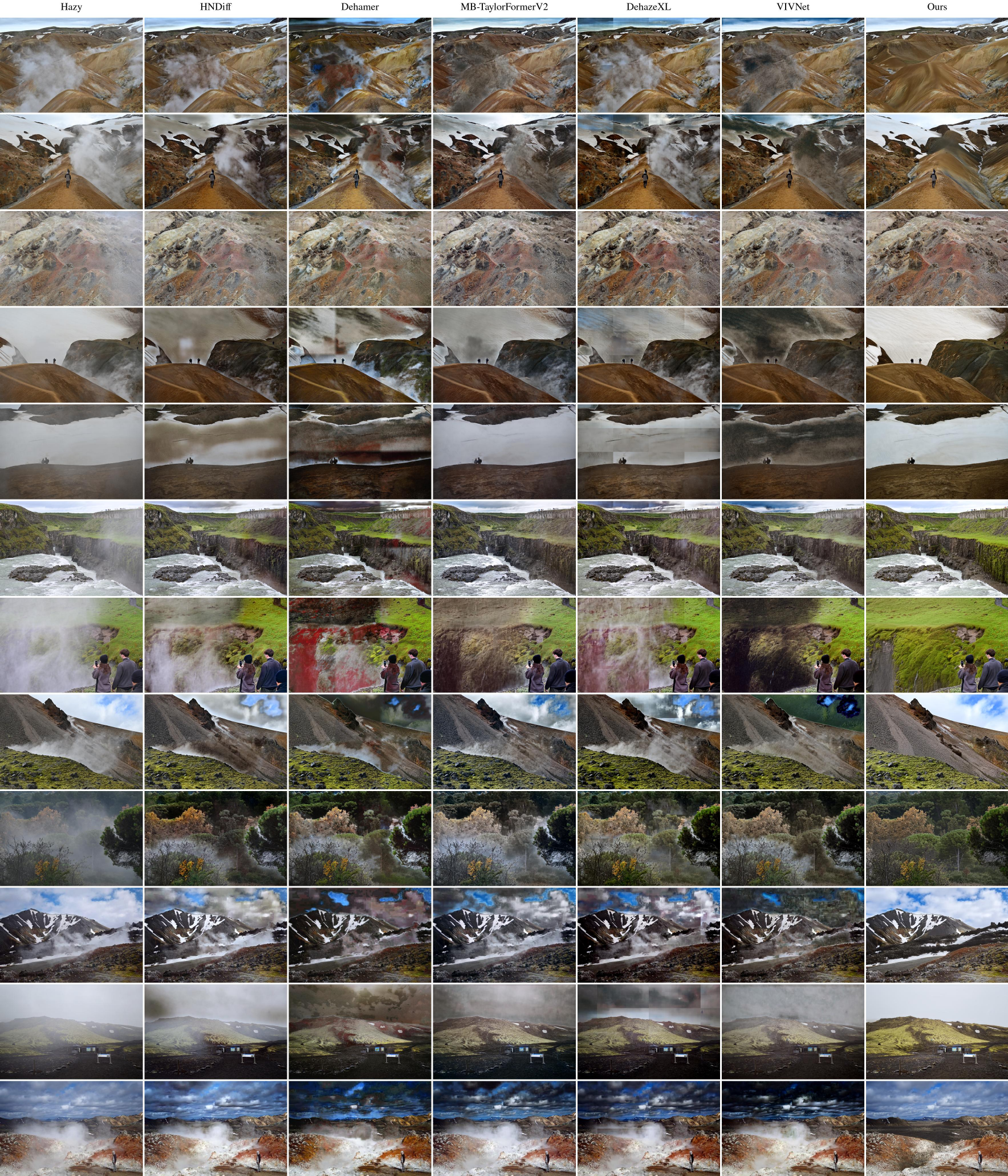}
  \caption{Examples of results in real hazy images. From left to right, we show input, HNDiff \cite{he2026hndiff}, DeHamer \cite{guo2022image}, MB-TaylorFormer V2 \cite{jin2025mb}, DehazeXL \cite{chen2025tokenize}, VIVNet \cite{cui2026visual}, and our method. Our method obtains more natural results, without any type of artifacts.}
  \label{fig:hazy 1}
\end{figure*}

\begin{figure*}[t!]
  \centering
  \includegraphics[width=\textwidth]{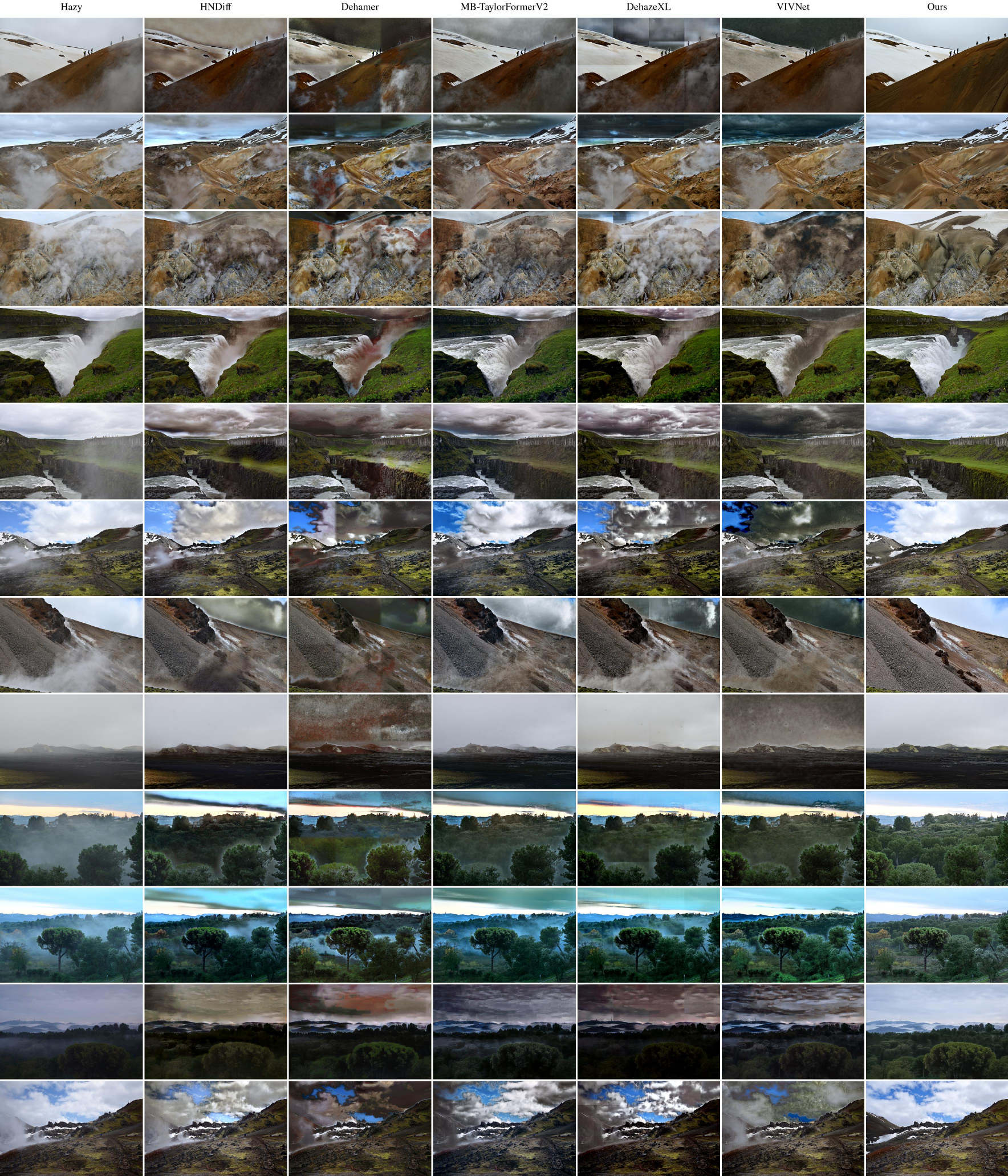}
  \caption{Further examples of results in real hazy images. From left to right, we show input, HNDiff \cite{he2026hndiff}, DeHamer \cite{guo2022image}, MB-TaylorFormer V2 \cite{jin2025mb}, DehazeXL \cite{chen2025tokenize}, VIVNet \cite{cui2026visual}, and our method. Our method obtains more natural results, without any type or artifacts.}
  \label{fig:hazy 2}
\end{figure*}

Fig.~\ref{fig:hazy 1} and Fig.~\ref{fig:hazy 2} show real hazy images processed by HNDiff \cite{he2026hndiff}, DeHamer \cite{guo2022image}, MB-TaylorFormer V2 \cite{jin2025mb}, DehazeXL \cite{chen2025tokenize}, VIVNet \cite{cui2026visual}, and our method. We observe the following
patterns. HNDiff, MB-TaylorFormer V2, DehazeXL, and VIVNet leave visible haze in many scenes (e.g., the hut and hillside, the steam on the slope, and the mountain valleys), while our results remove most of it. DeHamer produces strong color casts and patches of red or blue. VIVNet tends to darken the image and gives dark blotches in the sky and the haze regions. HNDiff often adds a brownish cast, while DehazeXL shows rectangular tile boundaries, and DeHamer shows similar seams.  In contrast, our outputs keep more natural colors in the same scenes, without introducing any artifacts. Furthermore, in dusk scenes, several baselines introduce sky banding, while our method does not.

\begin{table}[t!]
\centering
\caption{Ablation on LoRA ranks (Stage 1) on I-HAZE.}
\label{tab:ablation of rank}

\small
\setlength{\tabcolsep}{2pt}
\renewcommand{\arraystretch}{1}

\resizebox{\linewidth}{!}{
\begin{tabular}{lccccccccc}
\toprule
Rank
& PSNR$\uparrow$
& SSIM$\uparrow$
& LPIPS$\downarrow$
& $\Delta E\downarrow$
& FID$\downarrow$
& Q-Align$\uparrow$ \\
\midrule

$8$
& 23.95
& 0.845
& 0.441
& 5.410
& 59.61
& 4.443 \\

$16$
& 24.01
& 0.848
& \textbf{0.439 } 
& 5.282
& 55.17
& 4.456\\

$32~\text{(Ours)}$
& \textbf{24.55}
& \textbf{0.852}
& 0.463
& \textbf{5.275}
& \textbf{51.61}
& \textbf{4.458} \\

\bottomrule
\end{tabular}
}
\vspace{-2mm}
\end{table}

\begin{table}[t!]
\centering
\caption{Quantitative comparison on four real-world haze datasets.}
\label{tab:bilalora comparison}

\small
\setlength{\tabcolsep}{2pt}
\renewcommand{\arraystretch}{1.1}

\resizebox{\linewidth}{!}{
\begin{tabular}{lcccccccc}
\toprule
& \multicolumn{2}{c}{Dense-Haze}
& \multicolumn{2}{c}{I-Haze}
& \multicolumn{2}{c}{O-Haze}
& \multicolumn{2}{c}{NH-HAZE} \\
\cmidrule(lr){2-3}
\cmidrule(lr){4-5}
\cmidrule(lr){6-7}
\cmidrule(lr){8-9}

Model
& PSNR & SSIM
& PSNR & SSIM
& PSNR & SSIM
& PSNR & SSIM \\
\midrule

BiLaLoRA (released)
& 9.92 & 0.472
& 14.98 & 0.667
& 15.54 & 0.632
& 11.87 & 0.553 \\

BiLaLoRA (fine-tuned)
& 12.75 & 0.467
& 16.84 & 0.501
& 18.42 & 0.620
& 15.51 & 0.617 \\

Ours
& \textbf{18.01} & \textbf{0.628}
& \textbf{24.30} & \textbf{0.854}
& \textbf{26.46} & \textbf{0.797}
& \textbf{21.69} & \textbf{0.714} \\

\bottomrule
\end{tabular}
}
\end{table}

\begin{table}[t!]
\centering

\caption{Parameter count and inference speed on an RTX 4090.}
\label{tab:params-speed}
\resizebox{\linewidth}{!}{
\begin{tabular}{lcc}
\toprule
\textit{Parameters} & \multicolumn{2}{c}{Count} \\
\midrule
Stage 1: LoRA finetuning (trainable)  & \multicolumn{2}{c}{12.55M} \\
Stage 2: Gated image structure adapter (trainable)  & \multicolumn{2}{c}{33.4M}  \\
All (including Flux.2[klein], VAE)                          & \multicolumn{2}{c}{7.98B}  \\
\midrule
\textit{Inference time (RTX 4090)} & w/o Stage2 & w/ Stage2 \\
\midrule
$640\times512$               & 1.01\,s & 1.11\,s \\
$1024\times1024$             & 3.44\,s & 3.75\,s \\
\bottomrule
\end{tabular}}
\end{table}

\section{Inference Speed and Parameter Size}
\label{sup_sec:efficiency}
Table~\ref{tab:params-speed} reports the number of parameters and the inference time on a single RTX 4090. The trainable parameters are 12.55M for the Stage 1 LoRA adapters and 33.4M for the Stage 2 gated image structure adapter, for a total of about 46 M. The inference time for a 1024×1024 image is 3.75 s when processed by the whole model.


\clearpage

\end{document}